\documentclass[letterpaper, 10 pt, conference]{ieeeconf}  % Comment this line out if you need a4paper

\IEEEoverridecommandlockouts                              % This command is only needed if 
\usepackage{xcolor}
\usepackage{booktabs}
\usepackage{bm}
\usepackage{multirow}
\usepackage{multicol}
\usepackage{nicematrix}
\usepackage{threeparttable}
\usepackage{amssymb}
\usepackage{hyperref}
\usepackage{cleveref}
\let\labelindent\relax
\usepackage{caption}
\DeclareCaptionLabelSeparator{dottilde}{.~~}
\usepackage{enumitem}
\usepackage{subfigure}
\usepackage{algorithmic}

\author{Jingwen Yu$^{1,2}$, Hanjing Ye$^{2}$, Jianhao Jiao$^{3}$, Ping Tan$^{1}$ and Hong Zhang$^{2*}$
\thanks{*corresponding author: Hong Zhang {\tt\small hzhang@sustech.edu.cn}}
\thanks{$^{1}$The Hong Kong University of Science and Technology, Hong Kong SAR, China.
{\tt \small jingwen.yu@connect.ust.hk}}
\thanks{$^{2}$Shenzhen Key Laboratory of Robotics and Computer Vision, Southern University of
Science and Technology, Shenzhen, China.}
\thanks{$^{3}$Department of Computer Science, University College London, London, The United Kingdom.}
\thanks{The proposed benchmark is open-sourced at \url{https://github.com/jarvisyjw/GV-Bench}.}
}

\begin{document}
\title{\LARGE \bf
GV-Bench: Benchmarking Local Feature Matching for Geometric Verification of Long-term Loop Closure Detection}
\maketitle

\begin{abstract}
Visual loop closure detection is an important module in visual simultaneous localization and mapping (SLAM), which associates current camera observation with previously visited places.
Loop closures correct drifts in trajectory estimation to build a globally consistent map.
However, a false loop closure can be fatal, so verification is required as an additional step to ensure robustness by rejecting the false positive loops. 
Geometric verification has been a well-acknowledged solution that leverages spatial clues provided by local feature matching to find true positives.
Existing feature matching methods focus on homography and pose estimation in long-term visual localization, lacking references for geometric verification.
To fill the gap, this paper proposes a unified benchmark targeting geometric verification of loop closure detection under long-term conditional variations.
Furthermore, we evaluate six representative local feature matching methods (handcrafted and \textcolor{black}{learning-based}) under the benchmark, with in-depth analysis for limitations and future directions.
\end{abstract}
\section{Introduction}
The main objective of loop closure detection is to facilitate robust navigation for long-term and large-scale autonomy \cite{tsintotas2022revisiting}, where the robot's operation may span different day periods, weather conditions, and seasons. 
As a result, it becomes increasingly difficult to associate two images under  challenging conditional variations. Visual loop closure detection first extracts compact global descriptors from images \cite{liu2012visual}; then, the similarities are calculated, and results are retrieved via nearest neighbor searching.
Extensive works \cite{arandjelovic2016netvlad, berton2022rethinking, keetha2023anyloc, ye2023condition} focus on learning conditional-invariant global features under changing environments. 
A potential loop closure pair is formed by the current query image with a retrieved database image, as the retrieval stage in Fig. \ref{fig:pipeline}.
\begin{figure}[t]
    \centering
    \includegraphics[width=3.3in]{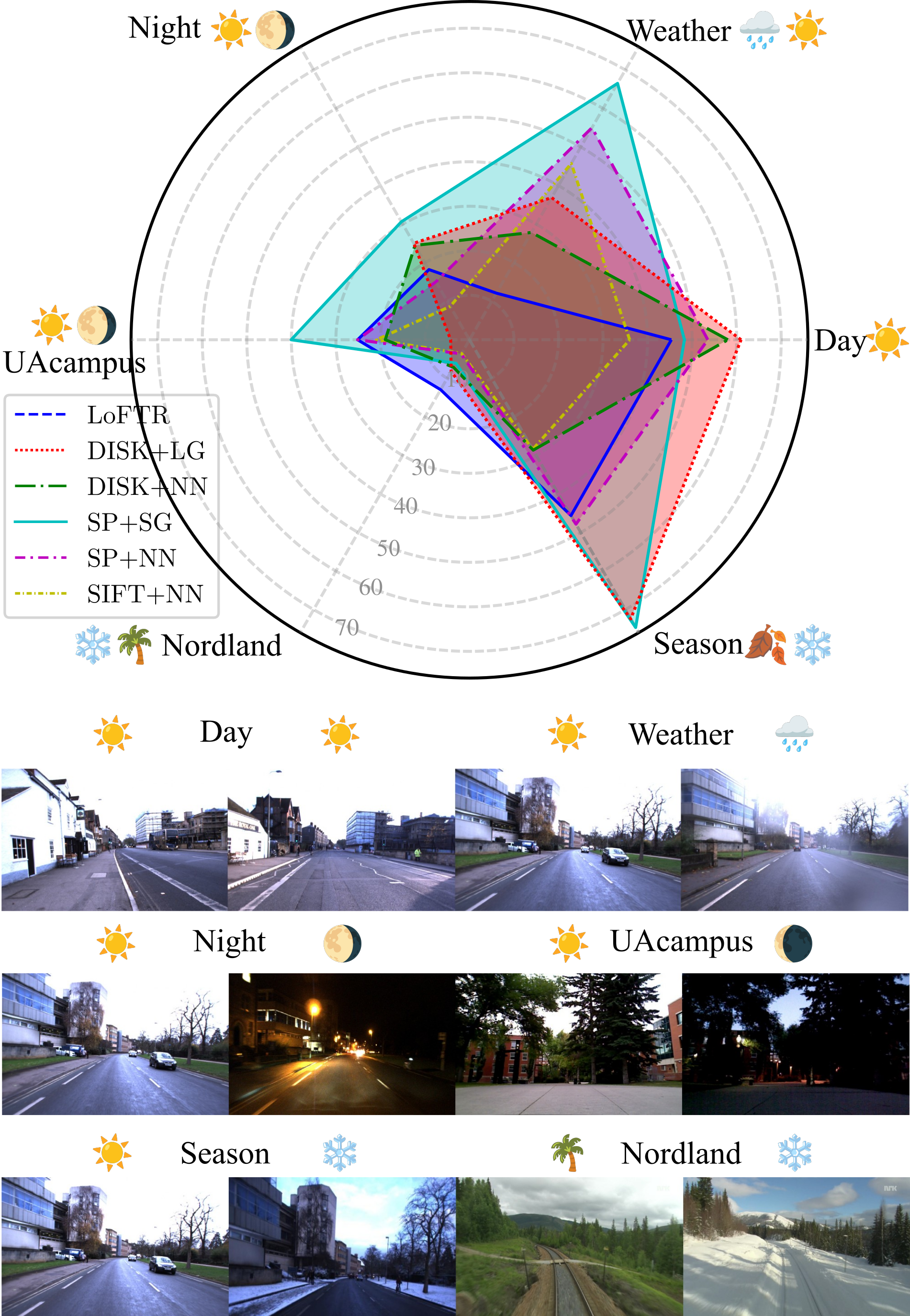}
    \caption{\textbf{GV-Bench} enables evaluation of geometric verification over long-term loop closure detection. The benchmark contains six sequences representing different conditional variations. The benchmark is built upon three datasets \cite{maddern20171, nordlanddataset, liu2015keypoint}. We carefully select and compare six representatives of local feature matching, where SP.+SG. exhibits the best performance on average. However, the unfilled and unbalanced radar chart indicates promising improvement. The metric used in the chart is max recall @100 precision, which is explained in Sec. \ref{sec:metrics}. The detailed construction of benchmark sequences is described in Sec. \ref{sec:benchmark intro}.}
    \label{fig:gvbench-overview}
\end{figure}
In detecting such a potential pair, only the appearance similarity between the images is used without considering the geometric relation between images. 
However, the appearance of places might change dramatically in the long term due to various conditions. Meanwhile, the appearance might be aliasing due to repetitive texture and similar structures.
Therefore, missing correct loop closures (i.e., low recall) and detecting wrong loop closures (i.e., low precision) are the main challenges.
To mitigate the second challenge, an additional step of geometric verification is employed to determine the correctness of each potential loop closure pair since false loop closures could lead to inaccurate localization.

Geometric verification refers to checking the spatial consistency of local feature correspondences between image pairs. 
As shown in Fig. \ref{fig:pipeline}, local features are first extracted and matched, followed by an outlier rejection method, such as RANSAC \cite{fischler1981random} by imposing epipolar constraint \cite{hartley2003multiple} between image pairs.
The number of inliers or the inlier ratio is commonly employed as a binary classifier to validate a potential loop closure pair.
A potential pair is classified as true if the number of inliers from local feature matching exceeds a pre-defined threshold. 
Therefore, conditional-invariant matches are the key to correctly verifying potential loop closure pairs.
The long-term visual localization benchmark \cite{toft2020long} provides a reference for choosing local features that are robust to appearance and viewpoint variations in pose estimation.
However, there is no benchmark for evaluating their performance on geometric verification.

To fill the gap, we propose a unified benchmark for evaluating local feature matching methods under geometric verification for long-term loop closure detection.
We aim to offer a reference for selecting a feature matching method for geometric verification in long-term autonomy.
Extensive experiments are conducted to compare six representative local feature matching methods, summarized in Table \ref{tab:matching-methods}. 
We designed our benchmark with an easy-to-extend framework on top of three datasets \cite{maddern20171, nordlanddataset, liu2015keypoint} targeting long-term visual loop closure detection and place recognition.
For a comparative study, we focus on three types of conditional variation: illumination, seasonal, and weather changes, as demonstrated in Fig. \ref{fig:gvbench-overview}.

In summary, the main contributions of this paper are:
\begin{itemize}
    \item \textbf{Fair and accessible geometric verification evaluation.} We open-source an out-of-box framework with a modular design as illustrated in Fig. \ref{fig:dataset}, allowing for evaluating newly proposed methods on the common ground and extending to more \textcolor{black}{diverse} datasets. 
    \item \textbf{A systematic analysis of geometric verification.} By employing the proposed benchmark, we
    point out possible future directions \textcolor{black}{(e.g., training feature extractor and matcher with conditional variation data) through extensive experiments.}
\end{itemize}
\begin{table}[t!]
    \centering
    \normalsize %\small \footnotesize
    \renewcommand\arraystretch{1}
    \caption{\textbf{Six selected representatives of local feature matching.} To simplify notation, we use the abbreviation below to represent each method.}
\begin{NiceTabular}{l@{\quad}l@{\quad}l@{\quad}}
\toprule
\textbf{Notation} & \textbf{Feature} & \textbf{Matcher} \\
\midrule
SIFT + NN & SIFT \cite{lowe1999object} & Nearest Neighbor\\
SP. + NN    &SuperPoint \cite{detone2018superpoint} & Nearest Neighbor \\
SP. + SG.  &SuperPoint \cite{detone2018superpoint} & SuperGlue \cite{sarlin2020superglue}   \\
DISK + NN  &DISK \cite{tyszkiewicz2020disk} & Nearest Neighbor \\
DISK + LG. &DISK \cite{tyszkiewicz2020disk}  & LightGlue \cite{lindenberger2023lightglue} \\
LoFTR      &          ---                  & LoFTR \cite{sun2021loftr} \\
\bottomrule
\end{NiceTabular}
    \label{tab:matching-methods}
\end{table}

\section{RELATED WORKS}
\label{sec:related}

\subsection{Loop Closure Verification}
\label{sec:related_verification}
Conventionally, geometric verification is a primary choice for verifying loop closure candidates \cite{tsintotas2022revisiting}.
Equivalently, it is also commonly adopted in visual localization systems, where it verifies the result from place recognition to help global localization \cite{sarlin2019coarse}. 
In addition to the spatial constraints between images, a series of works attempts to leverage keypoints topology \cite{yue2019robust}, scene graph \cite{hughes2022hydra}, and semantics \cite{orhan2022semantic}.
In the 3D reconstruction, similar structures of buildings may cause mistakes in structure-from-motion. In a recent work \cite{cai2023doppelgangers}, local feature matches combined with a convolutional neural network are employed as a binary classifier. However, it is limited by the domain gap and quality of image matching.
Though many verification methods exist, a unified benchmark for evaluation on the common ground is missing.
This paper proposes a benchmark to evaluate the geometric verification performance under long-term environmental changes, which are challenging for long-term robot autonomy. 
Besides, we specifically focus on evaluating different types of local feature matching methods 
under various conditions, aiming to reveal the existing challenges for future directions.
\begin{figure}[t]
    \centering
    \includegraphics[width=0.48\textwidth]{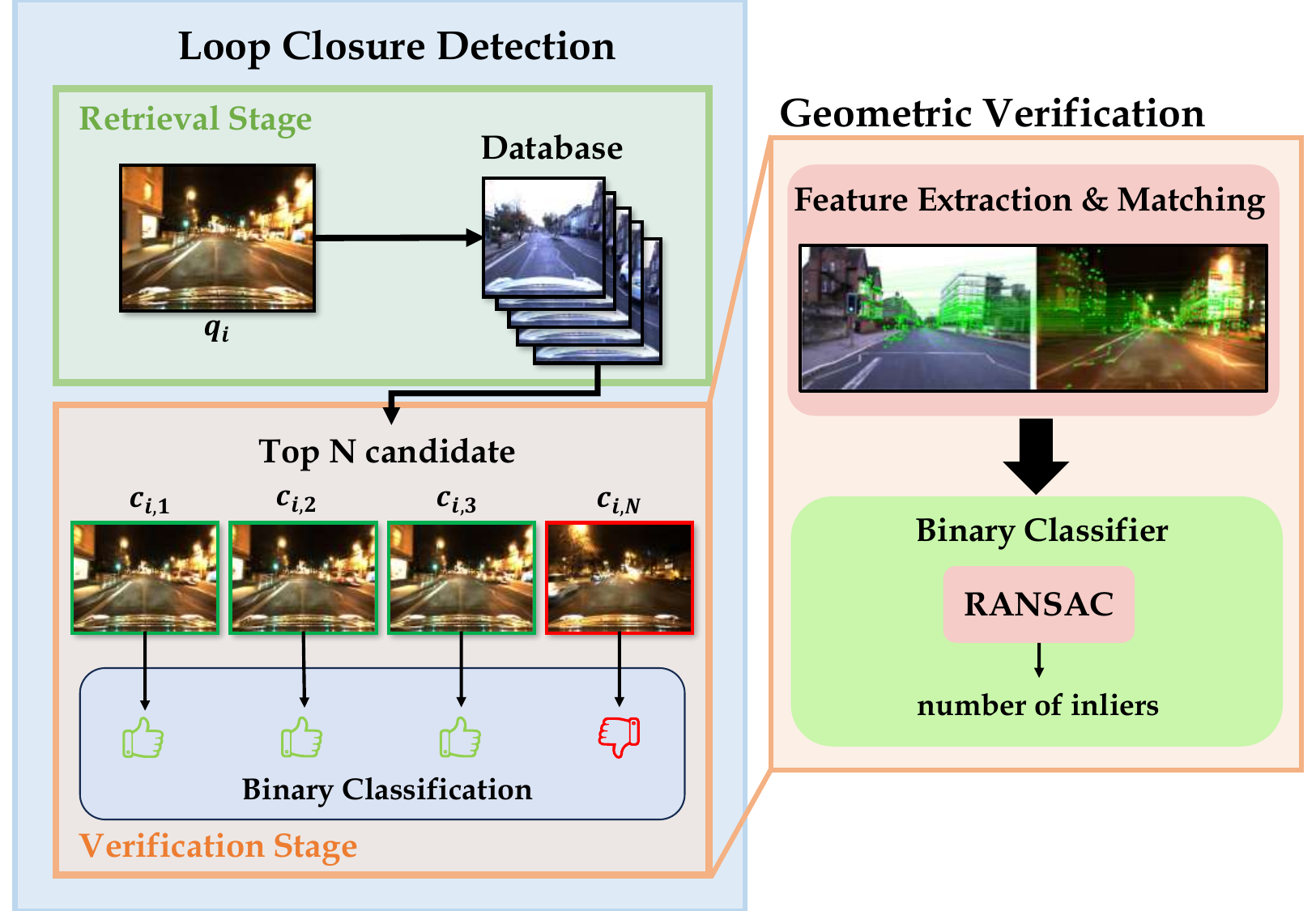}
    \caption{\textbf{Loop Closure Detection}
    consists of two stages: retrieval and verification. Potential loop closure pairs $\{q_{i}, c_{i,j}\}$ detected by the retrieval stage are sent for verification. Each pair of images is examined under geometric constraints provided by local feature matching. RANSAC filters the matched correspondences to find the best inliers, which is used as the probability in binary classification.}
    \label{fig:pipeline}
\end{figure}

\subsection{\textcolor{black}{Local Feature Matching}}
Local feature matching is \textcolor{black}{traditionally} performed by i) detecting keypoints and computing keypoint descriptors, ii) matching descriptors using a nearest neighbor (NN) search, iii) filtering outliers by RANSAC \cite{fischler1981random}. 
SIFT \cite{lowe1999object}, as a representative handcrafted feature, is widely used in object recognition, homography estimation, structure-from-motion, etc, for its scale and rotation invariance.
With the advent of convolutional neural networks (CNNs), data-driven methods focus on learning better sparse detectors and local descriptors, such as SuperPoint \cite{detone2018superpoint} and DISK \cite{tyszkiewicz2020disk}. SuperPoint learns keypoint detection and description jointly by self-supervision. DISK proposes learning keypoint detection using a probabilistic approach via reinforcement learning.

Inspired by the success of learning-based local features, SuperGlue \cite{sarlin2020superglue} achieves feature matching via a graph convolutional neural network (GCN), which performs particularly well with SuperPoint in homography and pose estimation. The following work, LightGlue \cite{lindenberger2023lightglue}, improves runtime performance and generalizes its usage to more feature keypoints and descriptors (including DISK). 
With the development of transformers, LoFTR \cite{sun2021loftr} predicts semi-dense matches and keypoints jointly, outperforming previous methods on pose estimation and claims to predict distinctive features under low-texture and repetitive patterns. 
In summary, we carefully select six representative local feature matching methods listed in Table \ref{tab:matching-methods} for evaluating geometric verification on the proposed benchmark.

\section{Evaluation Methodology}
\label{sec:eval}
\begin{figure}[ht]
    \centering
    \includegraphics[width=0.46\textwidth]{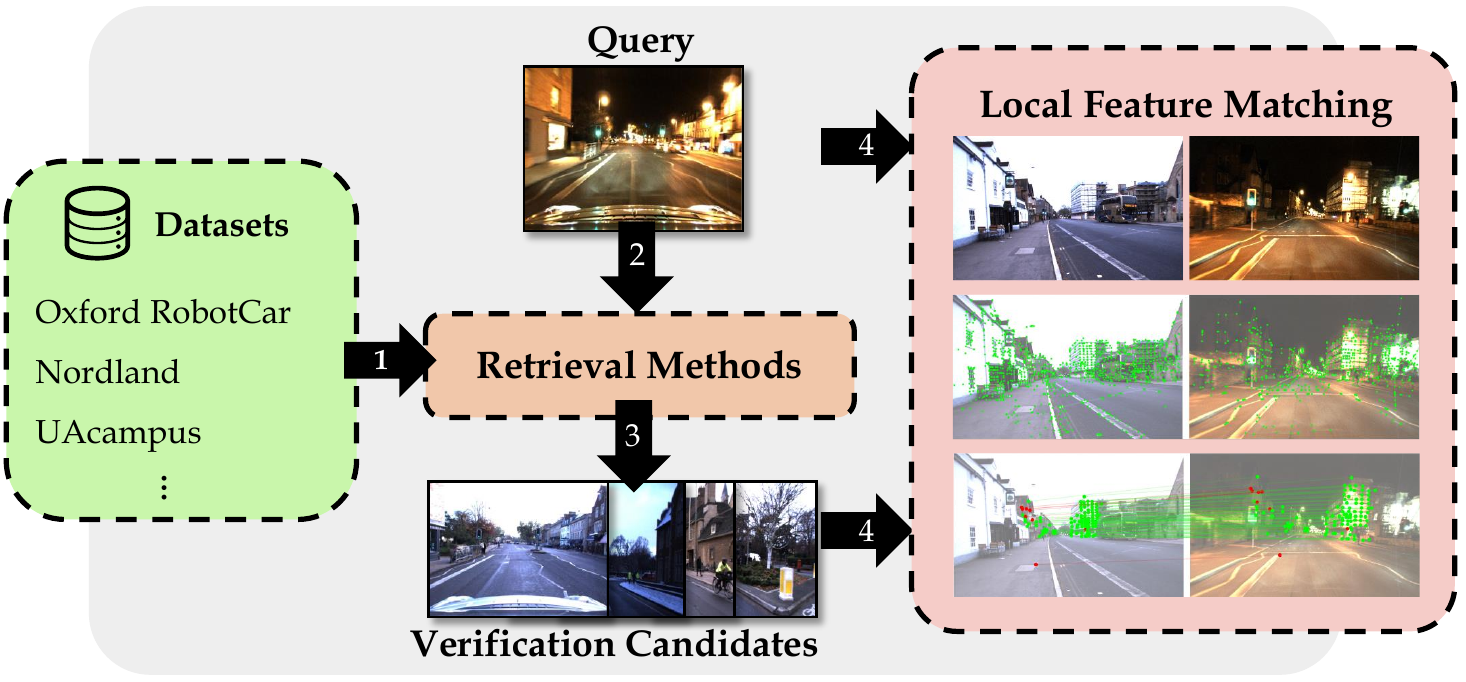}
    \caption{\textbf{The pipeline of open-sourced benchmark} consists of
    i) pre-process dataset, ii) randomly select query set (if the dataset does not provide it),
    iii) retrieve verification candidates for each query,
    iv) match queries with candidates.
    The dashed modules (Datasets, Retrieval Methods, and Local Feature Matching) are expendable in the open-sourced framework, enabling easy customization for research purposes (i.e., enlarging sequences, using other retrieval methods, and evaluating new feature matching methods.)
    }
    \label{fig:dataset}
\end{figure}

Here, we describe how a geometric verification method is typically constructed and how we evaluate different local feature matching algorithms.
In Fig. \ref{fig:pipeline}, geometric verification deals with an image pair, computing the probability that the given pair is positive (i.e., representing the same spatial location).
For each image pair, we first extract keypoints and compute correspondences from the feature matchers.
These matches are further processed by computing the fundamental matrix \cite{hartley2003multiple} assuming a non-planar scene and applying RANSAC \cite{fischler1981random} to reject outliers. 
The detailed steps are provided in Section \ref{sec:gv}.
Moreover, a consistent list of image pairs should be used for fair comparison.
In Section \ref{sec:benchmark intro}, we describe how we generate a unified benchmark for evaluating geometric verification subject to conditional and viewpoint variations using three datasets \cite{maddern20171, nordlanddataset, liu2015keypoint}. 
In Fig. \ref{fig:dataset}, we show the pipeline of benchmark construction, which is modular and easily expendable for more diverse conditional changes, local feature matching methods, and retrieval methods.

\subsection{Benchmark Introduction}
\label{sec:benchmark intro}
Geometric verification examines a potential loop closure pair consisting of the current or query frame $q_{i}$ with a database frame $r_{i}$. 
In loop closure detection, for the query frame $q_{i}$, we find the top $N$ candidates from the database by sorting the similarity scores. 
In our benchmark, we choose $N$ by the size of the image database.
Our benchmark generates image pairs by mimicking the retrieval stage of the loop closure detection. 
In other words, for each query frame $q_{i}$, a list of potential loop closure pairs is generated as:
\begin{equation}
\label{equ:pair}
     \{ \ q_{i}, \ r_{i,j} \ \}_{j = 1...N} \quad q_{i} \in \mathbf{Q},  \ r_{i,j} \in \mathbf{R}, \quad N \in \mathbb{R}  
     \end{equation}
where $\mathbf{Q}$ is the query set and $\mathbf{R}$ is the database set. We extract global descriptor $\mathbf{V}$ by the popular method NetVLAD \cite{arandjelovic2016netvlad}, which can be replaced by any retrieval method, thanks to the modular design illustrated in Fig \ref{fig:dataset}.

\textbf{Condition and viewpoint variations} are the main challenges of long-term loop closure detection, where robot traversals may span over day and night, different weather conditions, and even seasons. 
However, it is hard to quantify or control any variation. 
To our best effort, we leverage the following datasets: Oxford RobotCar dataset \cite{maddern20171}, Nordland dataset \cite{nordlanddataset}, and UAcampus \cite{liu2015keypoint} to control the variables of variations. 
In general, we design the benchmark with three controlled variations: illumination (day and night), seasonal, and weather variations, as shown in Table \ref{tab:dataset}. In addition to conditional variations, we build a baseline sequence ``Day'', for which only moderate environmental changes occur. Although this sequence is expected to be easily solvable, the results, as depicted in Fig. \ref{fig:pr_curve}, are suboptimal, indicating that further improvement is possible.

\subsubsection{Oxford Robotcar Dataset \cite{maddern20171}} The data was collected by mounted cameras on a vehicle in Oxford, UK. The dataset includes over 100 repetitions of a consistent route spanning over a year. 
This collection covers various seasons, weather, and times of the day, offering various environmental conditions. 
We derive four sub-sequences: ``Day'', ``Night'', ``Season'', and ``Weather'' from the original dataset as shown in Table \ref{tab:dataset}.
We randomly choose 1720 images from a sequence\footnote{2014-12-09-13-21-02} to form the query set $\mathbf{Q}$ for all sub-sequences.   
Then, we retrieve the top 20 candidates from the other four sequences\footnote{2014-11-18-13-20-12, 2014-12-16-18-44-24, 2015-02-03-08-45-10, 2014-11-25-09-18-32}.
The viewpoint changes are moderate since the images were captured on urban roads, as illustrated in Fig. \ref{sub:d} and \ref{sub:c}. 
While conditional changes strongly over day and night.
The ground truth labels for generated pairs are determined following the rule in \cite{warburg2020mapillary}, which selects positive pairs within a radius of 25 meters in translation and 40 degrees in orientation.
Reference poses for images are provided by the RTK measurements.
All the images are post-processed to avoid ambiguity\footnote{We undistort and remove the original images' car hood (bottom 160 rows of pixels).}.

\subsubsection{Nordland Dataset \cite{nordlanddataset}} The images were collected by a mounted camera on the Nordland Railway in Norway across seasons. Following \cite{camara2020visual}, we use images from winter as $\mathbf{Q}$ and summer as $\mathbf{R}$. 
We extract the top 20 candidates for 1415 queries to generate 28300 potential pairs.
The conditional variation is strong since the vegetation is covered in heavy snow, without artificial buildings as landmarks. 
However, the viewpoint change is approximated to none for a fixed camera posture on the train.
If the reference image is within two frames relative to the query, it is selected as a positive pair.
\subsubsection{UAcampus Dataset \cite{liu2015keypoint}} A forward-facing camera collected the images on a mobile robot teleoperated on a university campus covering the same route of 650 meters. The sequences are collected at different times (06:20 and 22:15) on the same day. We use the morning (06:20) as $\mathbf{Q}$ and the night (22:15) as $\mathbf{R}$, generating the top 10 candidates for 647 query images.
The conditional changes are very strong due to the illumination. 
The campus buildings are of similar appearance. In the meantime, the viewpoint variations are large. The ground truth is annotated manually.
\begin{table}[t]
\centering
\footnotesize
\caption{\textbf{Summary of sequences in proposed benchmark.} The benchmark consists of six sequences covering mainly three types of conditional changes: illumination (Night and UAcampus), seasonal (Season and Nordland), and weather changes in long-term loop closure detection. The ``Day'' sequence serves as the baseline challenge with moderate environmental changes over a short period.}
\renewcommand\arraystretch{1.2}
\begin{tabular}{lcccc}
\toprule
\multirow{2}{*}{\textbf{Sequences}} & \multicolumn{2}{c}{\textbf{Traverse}}    & \multicolumn{2}{c}{\textbf{Variations in }}   \\ 
\cmidrule(lr){2-3} 
\cmidrule(lr){4-5}
              & \multicolumn{1}{c}{\textbf{Query}} & \textbf{Ref.} & \multicolumn{1}{c}{\textbf{Condition}} &\textbf{Viewpoint}  \\
\midrule
Day \cite{maddern20171}& \multicolumn{1}{c}{Day} &Day & \multicolumn{1}{c}{moderate} &moderate   \\ 
Night \cite{maddern20171}    & \multicolumn{1}{c}{Day}     &Night   &\multicolumn{1}{c}{very strong} &moderate   \\ 
Season \cite{maddern20171}  & \multicolumn{1}{c}{Autumn}  &Winter  & \multicolumn{1}{c}{strong} &moderate  \\ 
Weather \cite{maddern20171}  & \multicolumn{1}{c}{Sun}  &Rain   & \multicolumn{1}{c}{strong} &moderate   \\ 
Nordland \cite{nordlanddataset} &\multicolumn{1}{c}{Winter}&Summer  & \multicolumn{1}{c}{very strong} &moderate   \\
UAcampus\cite{liu2015keypoint}   &\multicolumn{1}{c}{Day}&Night & \multicolumn{1}{c}{strong} &strong  \\ 
\bottomrule
\end{tabular}
\vspace{-0.05in}
\label{tab:dataset}
\end{table}

\begin{figure}[t]
	\centering
	\subfigure[Negative pair in ``Nordland'' with inliers: 9]
	{\label{sub:a}\centering\includegraphics[width=0.48\textwidth]{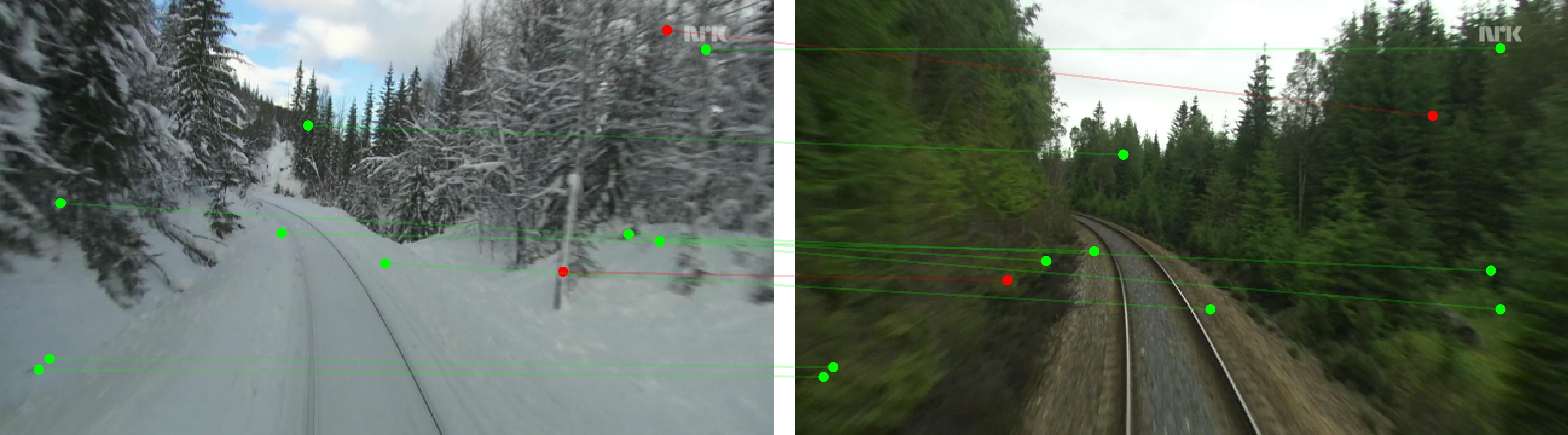}}
 \subfigure[Positive pair in ``Nordland'' with inliers: 37]
	{\label{sub:b}\centering\includegraphics[width=0.48\textwidth]{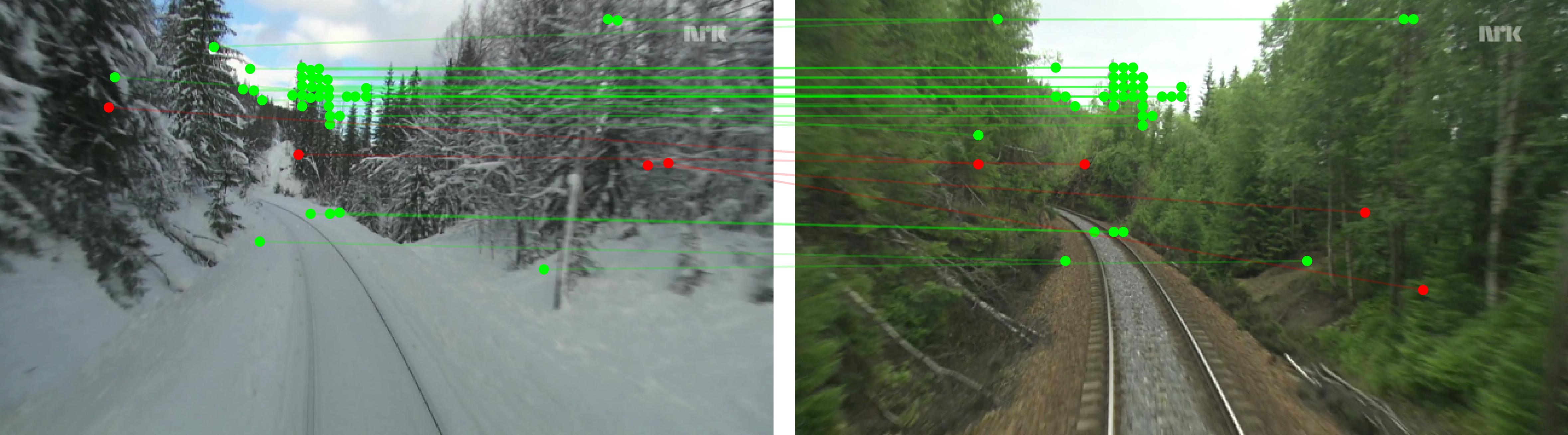}}
 \subfigure[Negative pair in ``Day'' with inliers: 45]{\label{sub:c}\centering\includegraphics[width=0.48\textwidth]{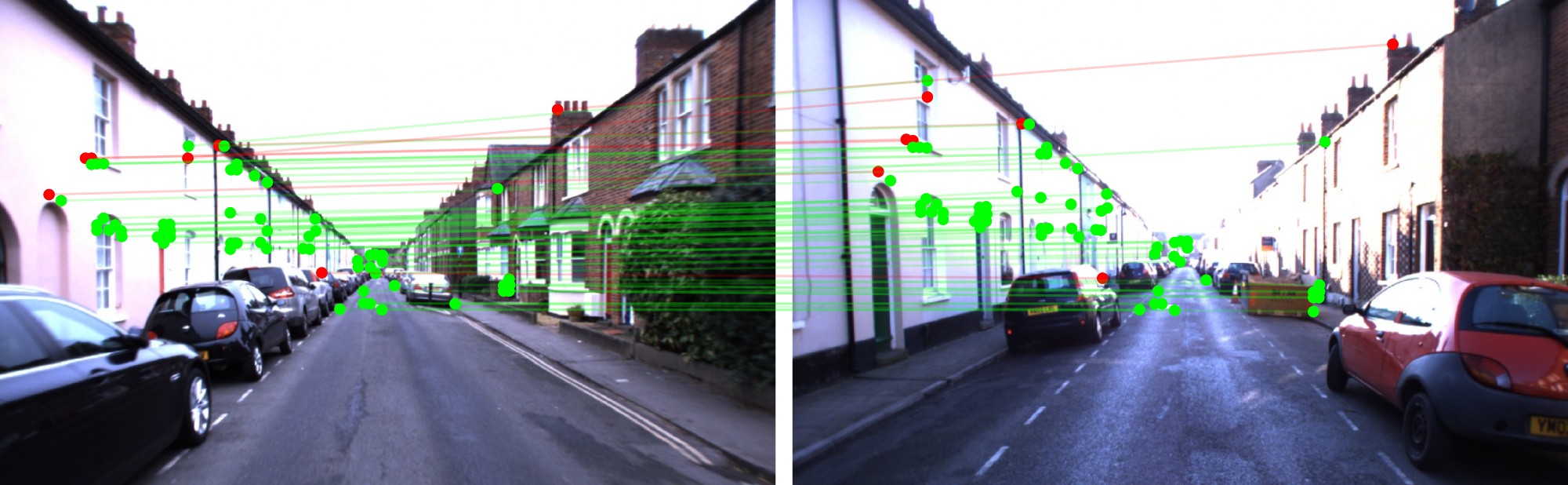}}
 \subfigure[Positive pair in ``Day'' with inliers: 41]{\label{sub:d}\centering\includegraphics[width=0.48\textwidth]{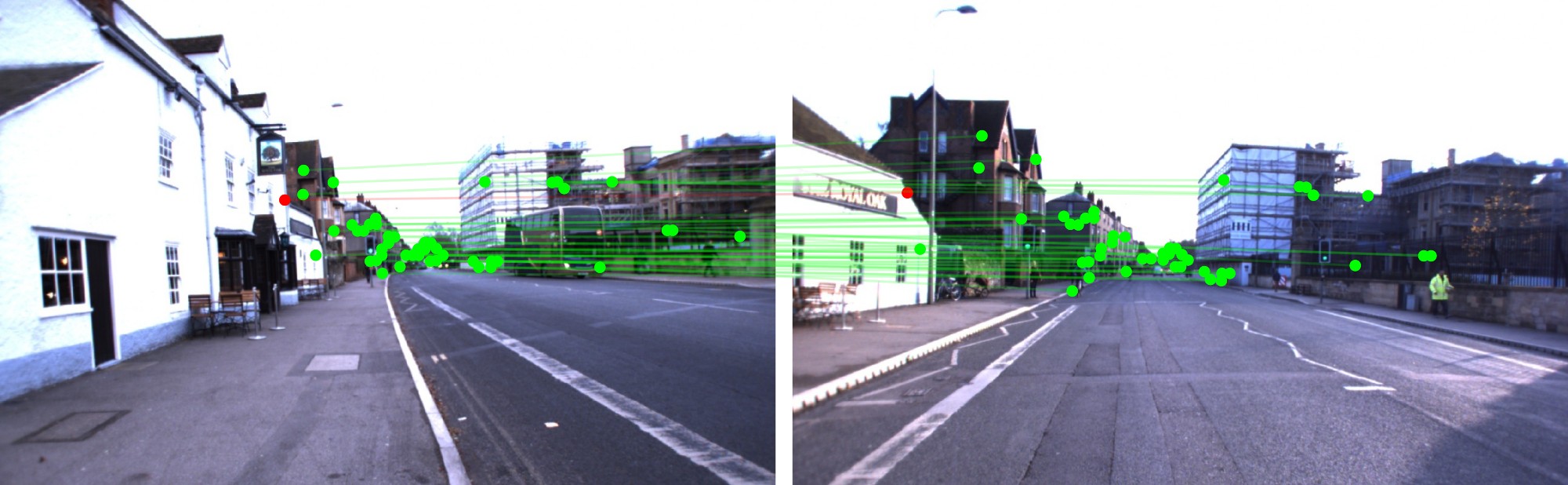}}
\caption{\textbf{Example of local feature matching} under long-term challenging conditional changes. In the above figures, we visualize LoFTR matches with vanilla RANSAC. The inliers (\textcolor{green}{green} lines) and outliers (\textcolor{red}{red} lines) are highlighted. The number of inliers of Fig. \ref{sub:c} and \ref{sub:d} are counter-intuitive because RANSAC fails when false matches are dominant (more detailed analysis are provided in Sec. \ref{sec:exp_diss}).}
 \label{fig:image-matching}
\end{figure}
%\footnotetext{https://github.com/jarvisyjw/GV-Bench}

\subsection{Geometric Verification}
\label{sec:gv}
\begin{figure}[ht]
    \centering
    \includegraphics[width=3.1in]{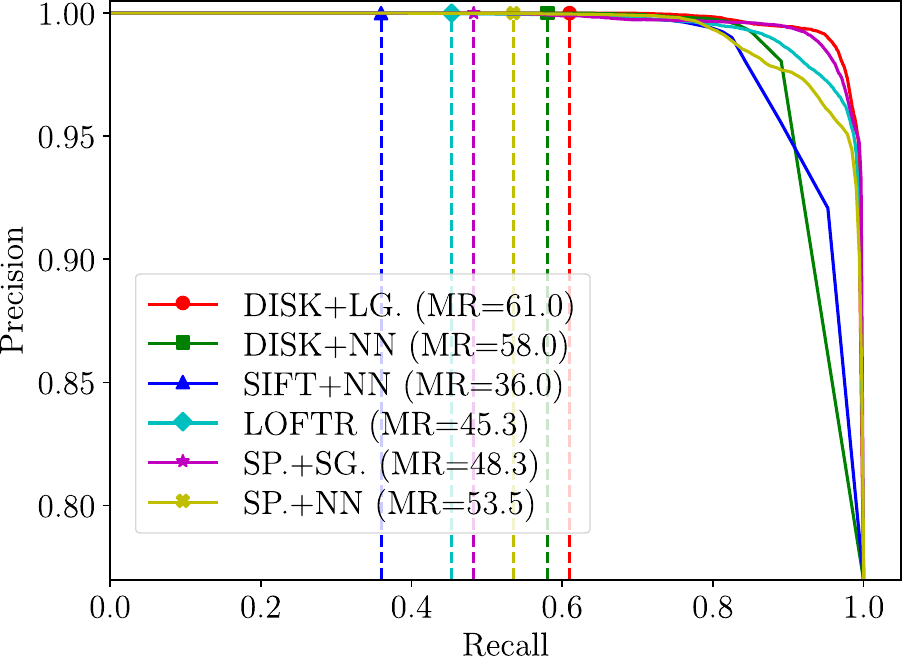}
    \caption{\textbf{Precision-recall curve} of the ``Day'' sequence. The marker annotates the maximum recall @100 precision (MR).}
    \label{fig:pr_curve}
    \vspace{-0.05in}
\end{figure}
Given an image pair, we extract keypoints and compute matches between them as demonstrated in Fig. \ref{fig:dataset}.
The matches are then transformed to point-wise correspondences denoted as $\mathbf{x_{1}}$ and $\mathbf{x_{2}}$ (in homogeneous coordinates), which are associated keypoint locations of the query and reference image, respectively.
\begin{equation}
\label{equ:funda}
\mathbf{x_{2}}^{\mathrm{T}} \mathbf{F} \mathbf{x_{1}}^{ } = 0
\end{equation}
We compute the fundamental matrix, denoted as $\mathbf{F}$, in conjunction with the RANdom SAmple Consensus (RANSAC) algorithm to solve for the best inliers $\mathcal{I}$. The fundamental matrix is a 3 by 3 matrix encapsulating the epipolar constraint between two views. It relates corresponding points in the two images through Equ. \ref{equ:funda}. The number of best inliers $\mathcal{I}$ is used for binary classification presented in Fig. \ref{fig:pipeline}. For example, if the number of $\mathcal{I}$ is larger than a given threshold, it is regarded as a positive pair.

\textbf{Six representatives of local feature matching methods} are carefully selected referring to long-term visual localization benchmarks \cite{toft2020long, Sattler_2018_CVPR}, and `Workshop on Long-Term Visual Localization under Changing Conditions\footnote{https://www.visuallocalization.net},' where local features are benchmarked for pose estimation under changing conditions. The combination of SuperPoint \cite{detone2018superpoint} and SuperGlue \cite{sarlin2020superglue} has been the lead for a long time till the detector-free local feature mather LoFTR \cite{sun2021loftr} came out. DISK \cite{tyszkiewicz2020disk} is chosen for its competitive performance demonstrated in Image Matching Challenge\footnote{https://www.cs.ubc.ca/research/image-matching-challenge/2020/} (IMW 2020). In addition to the learning-based methods, SIFT \cite{lowe1999object} is chosen to represent the classical handcrafted features. As listed in Table \ref{tab:matching-methods}, for SuperPoint and DISK, we compare the nearest neighbor (i.e., representing the distinctiveness of feature descriptor) and learning-based matching network for a comprehensive study.

\section{Experimental Results}
\label{sec:exp}

\begin{table*}[ht!]
\centering
\caption{\textbf{Experimental results of all six sequences.}
 \textbf{\textcolor{red}{Best}}, \textbf{\textcolor{blue}{Second Best}}, and \textbf{Third Best} results are highlighted. The last column presents the average performance of all six sequences. \textbf{MR} represents maximum recall @100 precision, \textbf{AP} stands for average precision. The results are round to three decimal places.}
	\renewcommand\arraystretch{1.20}
  \renewcommand\tabcolsep{5pt}
	\normalsize
 \begin{threeparttable}
     \begin{tabular}{lcccccccccccccc}
	\toprule[0.03cm]
 \multirow{2}{*}{\textbf{Matching Method}} & \multicolumn{2}{c}{\textbf{Day}} & \multicolumn{2}{c}{\textbf{Night}} & \multicolumn{2}{c}{\textbf{Season}} &\multicolumn{2}{c}{\textbf{Weather}} &\multicolumn{2}{c}{\textbf{Nordland}} &\multicolumn{2}{c}{\textbf{UAcampus}} &\multicolumn{2}{c}{\textbf{Average}}\\
   & \textbf{AP}  & \textbf{MR}  & \textbf{AP}  & \textbf{MR}  & \textbf{AP}  & \textbf{MR}  & \textbf{AP}  & \textbf{MR}  & \textbf{AP}  & \textbf{MR} & \textbf{AP}  & \textbf{MR} & \textbf{AP}  & \textbf{MR}\\
   \midrule[0.03cm]
    \textbf{SIFT + NN}   &98.1 &36.0 &62.1 &8.7  &98.9 &28.4   &99.7 &\textbf{45.7} &13.6 &3.5 &53.5 &\textbf{20.9} &71.0 &23.9 \\
    \textbf{SP. + NN}   &99.3 &\textbf{53.5} &\textbf{86.8} &15.1  &99.7 &\textbf{47.8}  &99.8 &\textbf{\textcolor{blue}{55.0}} &52.8 &3.7 &74.0 &\textbf{\textcolor{blue}{25.1}} &85.4 &\textbf{33.4}\\
    \textbf{SP. + SG.}   &\textbf{\textcolor{blue}{99.6}} &48.3 &\textbf{\textcolor{blue}{96.0}} &\textbf{\textcolor{red}{30.6}} &\textbf{\textcolor{red}{99.9}} &\textbf{\textcolor{red}{74.6}}&\textbf{\textcolor{red}{99.9}}&\textbf{\textcolor{red}{66.5}} &\textbf{\textcolor{blue}{71.1}}  &6.1 & \textbf{\textcolor{red}{85.3}}  & \textbf{\textcolor{red}{40.1}}
  &\textbf{\textcolor{blue}{92.0}}  &\textbf{\textcolor{red}{44.4}} \\
    \textbf{DISK + NN.} &97.4 &\textbf{\textcolor{blue}{58.0}} &51.7 &\textbf{24.5}  &99.0 &28.6 &99.7 &27.8 &17.9 &\textbf{7.2} &62.7  &18.9 &71.4 &27.5  \\
    \textbf{DISK + LG.}  &\textbf{\textcolor{red}{99.7}} &\textbf{\textcolor{red}{61.0}}  &81.4 &\textbf{\textcolor{blue}{25.0}}   &\textbf{99.9} &\textbf{\textcolor{blue}{72.4}}   &\textbf{\textcolor{blue}{99.9}} &36.9 &\textbf{63.8}&\textbf{\textcolor{blue}{8.2}} &\textbf{78.6} & 4.2 &\textbf{87.2} &\textbf{\textcolor{blue}{34.6}}\\
    \textbf{LoFTR}      &\textbf{99.5} &45.3 &\textbf{\textcolor{red}{97.9}} &18.2   &\textbf{\textcolor{blue}{99.9}} &45.6  &\textbf{99.9} &12.1  &\textbf{\textcolor{red}{81.1}} &\textbf{\textcolor{red}{13.0}} & \textbf{\textcolor{blue}{81.0}} & \textbf{\textcolor{blue}{25.1}} & \textbf{\textcolor{red}{93.2}}
    &26.6  \\
  \bottomrule[0.03cm]	
     \end{tabular}

 \end{threeparttable}

	\label{tab:exp-all}
\end{table*}

\subsection{Evaluation Metrics}
\label{sec:metrics}
Geometric verification can be regarded as a binary classification problem that determines whether a given potential loop closure pair is positive. 
Precision and recall are usually used for evaluation \cite{tsintotas2022revisiting}, where precision measures the proportion of correctly detected loop closures (true positive) out of all detected ones (true positives + false positives), and recall measures the proportion of correctly detected loop closures (true positives) out of all actual loop closures (true positives + false negatives).
High precision indicates a low rate of false alarms, while high recall indicates the ability to detect most of the true loop closures.
These two metrics are used jointly for the trade-off between false alarms and missed detections as a precision-recall curve in Fig. \ref{fig:pr_curve}. 
Geometric verification should reject the false positive loop closure robustly. 
Therefore, we use two metrics for evaluation: maximum recall @100 precision (\textbf{MR}) \cite{yue2019robust} and average precision (\textbf{AP}). 
The MR represents the highest recall while keeping the precision to 100{\%}, representing the ability to find true loop closures without false positives. 

\subsection{Comparison over Conditional Variations}
The evaluation results of the proposed benchmark are presented in Table \ref{tab:exp-all}.
On average, the combination of SuperPoint and SuperGlue achieves the highest MR of 44.4\%, while LoFTR reaches the highest AP of 93.2\%, indicating the highest AP does not lead to a higher MR.
The re-rank methods, like \cite{tan2021instance} and \cite{barbarani2023local} in the computer vision community, focus on evaluating the re-rank performance over AP (AUC), which does not provide a reference for geometric verification by far.
Although no method consistently ranks top on the benchmark under different challenging conditions, SP.+SG. achieves the highest MR on four sequences out of six, suggesting a decent choice for geometric verification.
LoFTR and DISK+LG. take up the remaining two first places, which are also competitive. 
DISK+LG. gets second-best MR on average; however, its poor performance on ``UAcampus'' indicates it might be unreliable.
Learning-based sparse feature extractors SuperPoint and DISK outperform SIFT by nearest neighbor matching, demonstrating the distinctiveness of learning-based descriptors under changing conditions. This is consistent with previous evaluations on homography \cite{detone2018superpoint} and pose estimation \cite{tyszkiewicz2020disk}.

\begin{figure}[ht]
\centering
\subfigure{\includegraphics[width=0.35\textwidth] {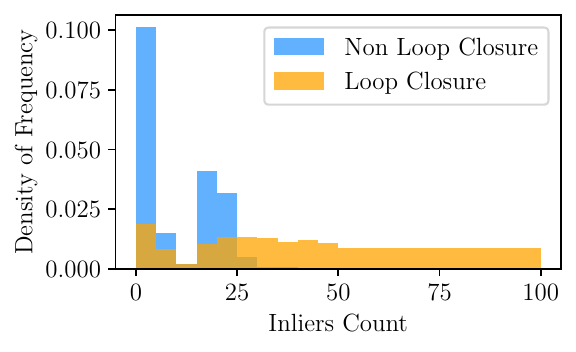}} \\
\vspace{-0.1in}
\subfigure{\includegraphics[width=0.35\textwidth]{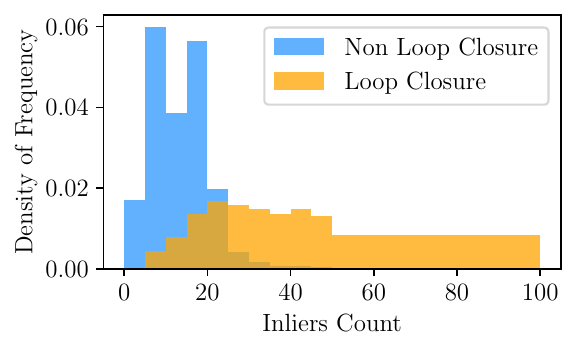}}
\vspace{-0.05in}
\caption{\textbf{The number of inliers' distribution} of the sequence ``Nordland'' of SP. + SG. (Top) and LoFTR (Bottom). The density of frequency indicates the distribution of the image pairs with the corresponding normalized inliers count. The ideal distribution is that there is no overlap between non-loop-closure pairs and loop-closure pairs. However, as figures above shows, it is rare in practice. From the plot of inliers' distribution, we can observe which kind of samples are in dominant.}
\vspace{-0.15in}
\label{fig:inliers_dist}
\end{figure}

\textbf{All methods suffer from perceptual aliasing.} As depicted in Fig. \ref{sub:a} and \ref{sub:b}, in ``Nordland'' sequences, the surrounding is unstructured vegetation. Even for loop closure pairs, the number of inlier matches is low, making it hard to distinguish from the negative pairs. Fig. \ref{fig:inliers_dist} demonstrated that the frequency density for LoFTR matches is more ``distributed,'' i.e., the average number of inliers of loop closures is higher than non-loop closures. The winter images are similar to a low-texture environment where the heavy white snow covers the rich texture of the environment. Our results are consistent with \cite{he2023detector} that LoFTR works especially well under texture-less situations. 

\subsection{\textcolor{black}{Comparsion over Runtime}}
\begin{figure}[ht]
\centering
\includegraphics[width=0.45\textwidth]{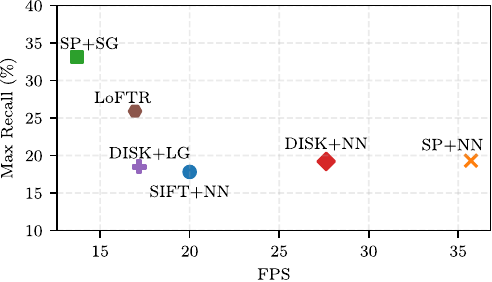}
\caption{\textbf{The efficiency vs. efficacy} of the six feature matching methods. Y-axis represents for performance in max recall @100 precision, x-axis is the inference speed in FPS.}
\label{fig:runtime}
\vspace{-0.1in}
\end{figure}

For robotics applications, time efficiency is critical on resource-constraint platforms.
We measure the runtime of six methods listed in Table \ref{tab:matching-methods} on NVIDIA GeForce RTX 3090 GPU and Intel i7-13700K CPU over 10K runs. 
The results are shown in Fig. \ref{fig:runtime} as inference time over performance, i.e., max recall @100 precision.
We can conclude that the runtime of six local feature matching methods is at a millisecond level on a modern GPU.
The choice can be made based on the trade-off between time efficiency and performance.

\subsection{Discussion}
\label{sec:exp_diss}
Based on the experiments, we conclude three key takeaways for the potential improvement of geometric verification: improving local feature extraction and matching, building a multi-condition image database, and enhancing outlier rejection.

\subsubsection{\textbf{Leveraging conditional variation data}}
Although feature extraction and matching methods like SP.+SG. and LoFTR perform well in long-term visual localization \cite{Sattler_2018_CVPR}, they do not work very well in our proposed benchmark. One of the possible solutions is to learn more distinctive descriptors under various conditions. SuperPoint is trained on synthetic data and MS-COCO \cite{lin2014microsoft}, while LoFTR is trained from MegaDepth \cite{MegaDepthLi18}. These datasets are not designed for long-term visual tasks and do not cover challenging conditional variations presented in our proposed benchmark.
A natural idea is to expand the diversity of the training set, i.e., training the feature detector and matcher with conditional variation data.
A recent work GIM \cite{xuelun2024gim}, shows the possibility of enhancing the robustness against long-term variations via learning generalizable image
matching from large amounts of data, such as internet videos.

Moreover, in the training of feature matchers \cite{sarlin2020superglue, lindenberger2023lightglue, sun2021loftr}, the goal is to match two paired images for homography or pose estimation. Therefore, a visual overlap is assumed, which is not hypothetical in geometric verification. Unpaired image samples should be considered in the training of feature matchers for robust geometric verification.

\subsubsection{\textbf{Building a multi-condition image database}}
Reducing the appearance gap between query and reference set may potentially mitigate the issue of perceptual aliasing illustrated in Fig. \ref{sub:a} and \ref{sub:b}.
According to a robotic visual localization system across long-term operations \cite{paton2016bridging}, multiple experiences (i.e., image sequences over different conditions) are stored in the database. 
It achieves robust localization under significant appearance variation by searching and matching the latest sequences. 
\textcolor{black}{As an inspiration, storing multiple sequences based on conditional changes (e.g., illumination, season, and weather changes) might alleviate the perceptual aliasing due to appearance gap over a long period.}

\subsubsection{\textbf{Exploiting more powerful outlier rejection}}
The vanilla RANSAC fails to work for scenarios where false matches dominate. As illustrated in Fig. \ref{sub:c}, in an urban environment with a similar structure, the incorrect matches are uniformly distributed, leading to a ``false fundamental matrix''. This happens to all the sequences regardless of conditional variation. 
RANSAC works under epipolar constraints only, which does not hold for two unpaired images. 
However, more powerful outlier rejectors \cite{wei2023generalized, barath2020magsac++} outperform in fundamental matrix estimation, promising possible improvement in geometric verification.

\section{Conclusions and Future work}
\label{sec:con}
This paper builds a unified benchmark for evaluating geometric verification of long-term visual loop closure detection. 
Specifically, we focus on conducting a comparative study of six carefully selected feature matching methods, covering handcrafted and learning-based features. 
Our benchmark is open-sourced with a modular design to support future research. 
This study allows us to draw the following conclusions. 
i) The combination of sparse keypoint detector SuperPoint \cite{detone2018superpoint} and feature matcher SuperGlue \cite{sarlin2020superglue} in general outperforms others.
ii) When subjected to challenging conditions, the learning-based keypoint descriptor is more distinctive than the handcrafted one. This result is consistent with previous evaluations on homography and pose estimation \cite{detone2018superpoint, tyszkiewicz2020disk}. 
iii) We conclude that the verification of loop closure detection is far from being solved. 
Further discussion suggests possible future directions, including training keypoint detectors with conditional variation data, expanding the dataset with images collected under different conditions, and exploring more powerful outlier rejection methods. 

For future work, i) we will expand the benchmark to support other loop closure verification methods mentioned in Sec. \ref{sec:related_verification}. Although we designed this benchmark to focus on geometric verification, it naturally supports evaluating other verification methods. By this expansion, we can comprehensively understand and explore beyond geometric verification. 
ii) we will focus on the directions proposed in Sec. \ref{sec:exp_diss}, aiming at developing a novel and effective method of geometric verification.

\section*{ACKNOWLEDGMENT}
The authors appreciate the discussion with Mr. Guangcheng Chen (Southern University of Science and Technology), Mr. Zhonghang Liu (Singapore Management University), Dr. Hengli Wang (Huawei Technologies Co., Ltd), and Dr. Yuxuan Liu (TIER IV).

\bibliographystyle{IEEEtran}
\bibliography{IEEEabrv,ref}

\newpage
\section{Supplementary Material}
In this section, we look into the details of the experiments and benchmark, hoping to resolve issues encountered in geometric verification (GV).
\begin{table*}[ht!]
\centering
\caption{\textbf{Experimental results of all six sequences.}
 \textbf{Best} results are highlighted. The last column presents the average performance of all six sequences. \textbf{MR} represents maximum recall @100 precision, \textbf{AP} stands for average precision.}
	\renewcommand\arraystretch{1.20}
  \renewcommand\tabcolsep{5pt}
	\normalsize
 \begin{threeparttable}
     \begin{tabular}{lcccccccccccccc}
	\toprule[0.03cm]
 \multirow{2}{*}{\textbf{Matching Method}} & \multicolumn{2}{c}{\textbf{Day}} & \multicolumn{2}{c}{\textbf{Night}} & \multicolumn{2}{c}{\textbf{Season}} &\multicolumn{2}{c}{\textbf{Weather}} &\multicolumn{2}{c}{\textbf{Nordland}} &\multicolumn{2}{c}{\textbf{UAcampus}} &\multicolumn{2}{c}{\textbf{Average}}\\
   & \textbf{AP}  & \textbf{MR}  & \textbf{AP}  & \textbf{MR}  & \textbf{AP}  & \textbf{MR}  & \textbf{AP}  & \textbf{MR}  & \textbf{AP}  & \textbf{MR} & \textbf{AP}  & \textbf{MR} & \textbf{AP}  & \textbf{MR}\\
   \midrule[0.03cm]
    \textbf{Doppelgangers\cite{cai2023doppelgangers}}   &97.1 &35.5 &60.8 &2.0 &99.1&22.4 &99.6 &30.0 &65.2 &2.1 &30.1 &13.2     &75.3 &17.5 \\
    \textbf{(GV) SP. + SG.}   &\textbf{99.6} &\textbf{48.3} &96.0 &\textbf{30.6} &\textbf{99.9} &\textbf{74.6} &\textbf{99.9} &\textbf{66.5}  &71.1   &6.1   &\textbf{85.3}  &\textbf{40.1}  & 92.0  &\textbf{44.4}  \\
    \textbf{(GV) LoFTR}      &99.5 &45.3 &\textbf{97.9}&18.2 &97.9 &18.2 &99.9 &12.1  &\textbf{81.1}  &\textbf{13.0}    &81.0 &25.1 &\textbf{93.2} &26.6  \\
  \bottomrule[0.03cm]	
     \end{tabular}
 \end{threeparttable}
	\label{tab:exp-all}
\end{table*}
\begin{figure}[ht]
\centering
\includegraphics[width=0.35\textwidth]{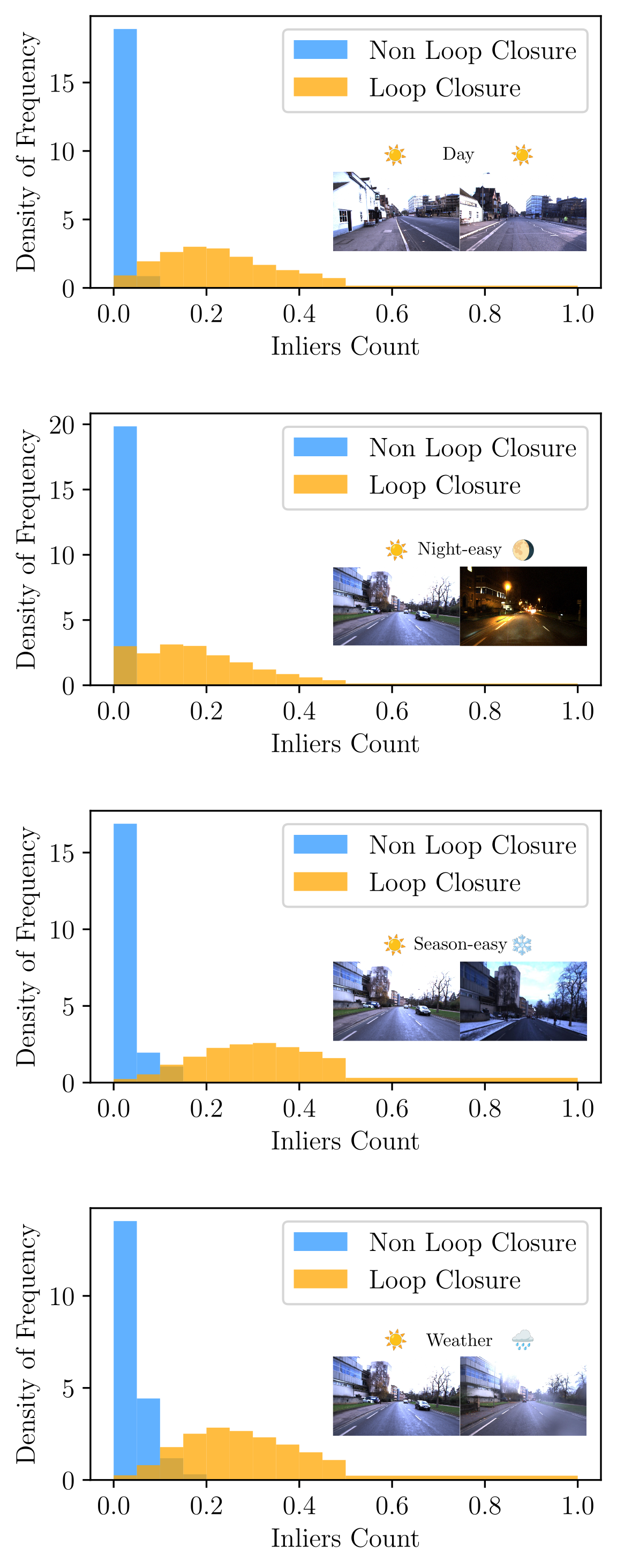}
\caption{\textbf{The number of inliers' distribution} of SP. + SG. on four sequences. The density of frequency indicates the distribution of the image pairs with the normalized inliers count. The ideal distribution is that there is no overlap between non-loop-closures and loop-closures. However, as figures above shows, it is rare in practice. There are two kinds of ``disturbance'' in the verification process, FPs (false-positives) and FNs (false-negatives). In the overlapped areas, non-loop-closures with non-zero matches are FPs, and loop-closures with zero matches are FNs.}
\label{fig:sp-sg-easy}
\end{figure}

\subsection{Matching Statistics}
\begin{figure}[ht]
\centering
\includegraphics[width=0.5\textwidth]{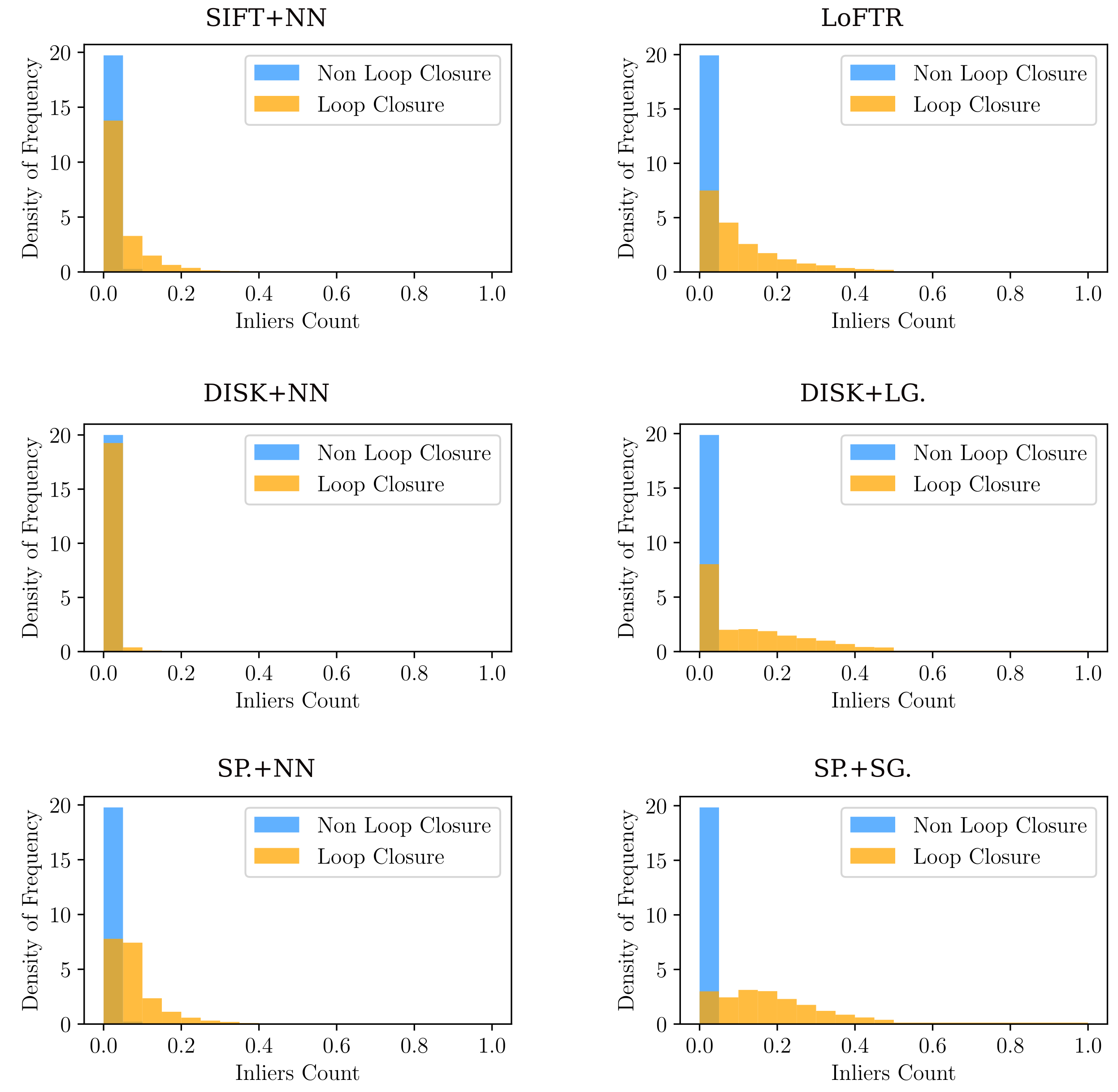}
\caption{\textbf{The number of inliers' distribution} of all methods on the Night sequence. In the Night sequence, the large illumination change results in zero matches for loop-closures, this is the major issue causing the low recall rate at 100 precision. }
\label{fig:night-easy-all}
\end{figure}

In this section, we provide more statistics about number of matches. In Fig. \ref{fig:sp-sg-easy}, the inliers' distributions on the Day, Night, Season, and Weather sequences are shown. Roughly speaking, in the Day, Season, and Weather sequences, the distribution of loop closures is somewhere between a ``distorted'' gaussian distribution. In the Night sequence, the shape of the distribution is different, where the distribution of loop closures with 0 matches increases. This is the main reason why the Night sequence is hard. Due to the large illumination changes, the local features might differ significantly. 
We can observe similar results in Fig. \ref{fig:night-easy-all}. All six methods suffer from this ``no matching'' challenge under significant illumination changes in the Night sequence. Among all the methods, the combination of SuperPoint and SuperGlue achieves the best performance by tending to generate more matches. Attention-based matching networks (SuperGlue and LightGLue) can alleviate this suffering, as shown in the comparison between DISK+NN and DISK+LG and SP+NN and SP+SG. Therefore, developing a powerful image matching method could be the answer to geometric verification. 

\vspace{-0.1in}
\subsection{The Ambiguity in Ground Truth}
\begin{figure}
    \centering
    \includegraphics[width=0.48\textwidth]{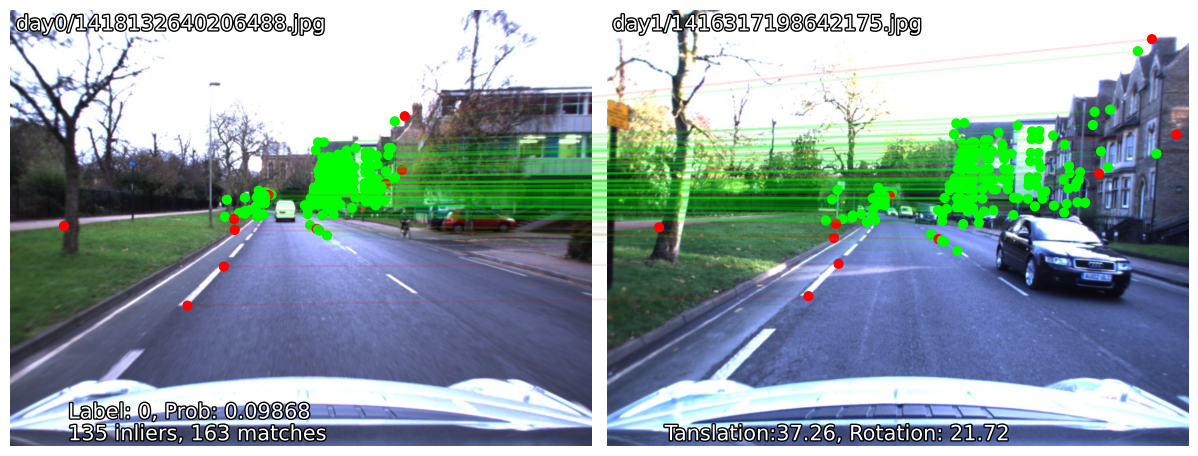}
    \caption{A ambiguity sample in Day sequence with SP.+SG. matches. The inliers (\textcolor{green}{green} lines) and outliers (\textcolor{red}{red} lines) are highlighted. According to the GPS, the distance between two images is 37.26m / 21.72$^{o}$. This distance is slightly over the threshold 25m / 40${^o}$, while the matches are correct since the urban street is in parallel with the camera's viewing direction.}
    \label{fig:gt-label}
\end{figure}

\begin{figure}
    \centering
    \includegraphics[width=0.42\textwidth]{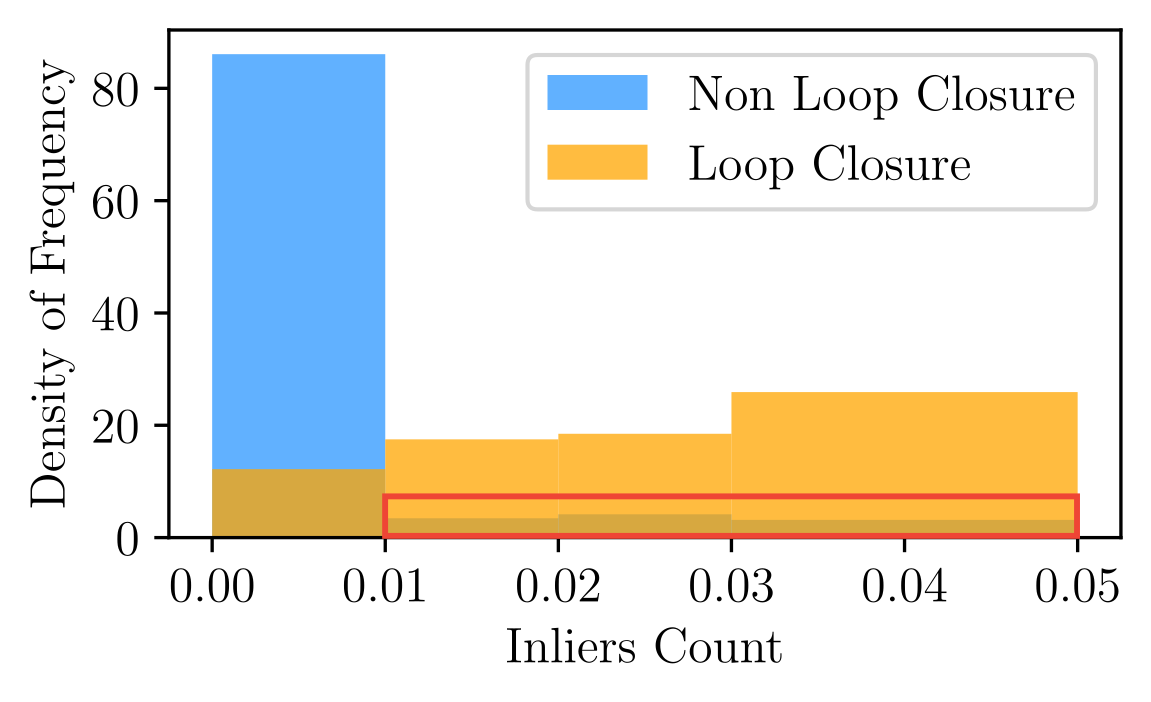}
    \caption{\textbf{The number of inliers' distribution} of all SP.+SG. on Day sequence with a zoom-in display. The area within the red bounding box indicates the non-loop-closures within the overlapped area between two distributions. The non-loop-closures get non-zero matches, causing the recall rate @ 100 precision to be low. As we sample in this area, finding that the dominant cases are shown in Fig. \ref{fig:gt-label}, where the ground truth definition might be ambiguous by following the existing criteria in \cite{warburg2020mapillary} and \cite{arandjelovic2016netvlad}.}
    \label{fig:gt-label-stat}
\end{figure}

In the current version of the benchmark, the ground truth label generation follows the general rule in most existing outdoor VPR methods \cite{arandjelovic2016netvlad, warburg2020mapillary}, where a hard threshold on the GPS measurements is employed. However, as we dig further into the false positives, we find that ambiguities exist in this labeling strategy. In Fig. \ref{fig:gt-label}, the distance between two images exceeds the threshold (25m / 40${^o}$), but a distinctive and feature-rich building exists in the scene, which is informative for downstream tasks of VPR, e.g., pose estimation. Since GV-Bench follows the existing standard for ground truth labeling, this kind of samples are considered negative. The conclusion is drawn by randomly sampling from the top 100 false positives of the Day sequence, as shown in the red box in Fig. \ref{fig:gt-label-stat}. This is a tricky issue that has not drawn researchers' attention in recent years, which might be trivial for developing a VPR system. However, we argue its importance in the downstream tasks, especially in AR/VR and Robotics applications. The ground truth should be labeled favoring the designed tasks, e.g., pose estimation. The potential solutions include designing a dataset-specific threshold and defining a new way of labeling. In the existing works \cite{wei2024breaking, ge2020self}, the idea of using local geometrics in VPR has been emphasized. Leveraging local geometrics to develop a self-supervised scheme is a promising direction to alleviate this ambiguity for localization tasks.

\subsection{Beyond Geometric Verification}
In GV-Bench, we focus on benchmarking different local features for geometric verification, while there are other methods that can be used to distinguish whether two visual measurements represent the same spatial location. In Sec. \ref{sec:related_verification}, we reviewed several methods for loop closure verification, here we expand our experiments in comparison to geometric verification (GV). And since 
\subsubsection{Learning-based Method.}
Doppelgangers \cite{cai2023doppelgangers} trains a CNN to be a binary classifier with the input of a pair of RGB images and masks of their LoFTR \cite{sun2021loftr} matches. The output is the positive probability that the pair of images are of the same spatial location and viewing direction. 
It is originally designed to disambiguate views of symmetrical historic buildings for Structure-from-Motion.
Therefore, its idea is formulated as a ``spot-the-difference'' challenge, where the distribution of the semi-dense matches is employed as a complementary to RGB images for better performance.
Intuitively, the performance of Doppelgangers partially depends on the performance of the local feature extraction and matching method. 
Since currently, the training set of the benchmark is missing, we test the pre-trained model of Doppelgangers (pre-trained on MegaDepth \cite{MegaDepthLi18} and Doppelgangers Dataset \cite{cai2023doppelgangers}). The Doppelgangers perform poorly on GV-Bench without any fine-tuning, with notably bad results on the Night and Nordland sequences. Potential reasons might be the domain gap between the training set of Doppelgangers and GV-Bench. In the future work of GV-Bench, an in-domain training set should be provided to support the development of learning-based methods. 
\subsubsection{Uncertainty Estimation for Visual Place Recognition}
Recent work SUE \cite{zaffar2024estimation} models the task of verification as uncertainty estimation in visual place recognition (VPR). 
VPR and LCD share a similar retrieval stage. Therefore, the verification problem in LCD equivalently exists in VPR systems. 
SUE leverages the spatial localization of the database images to estimate the positive probabilities of K retrieval candidates. 
Its experiments prove that it performs better than learning-based uncertainty estimation methods while falling behind GV. 
However, SUE argues that the time consumed for GV is much longer. 
In the future expansion of GV-Bench, SUE should be considered as a baseline for loop closure verification.
In a SLAM system, the relative poses of the database images are free-of-charge and often neglected.

\end{document}